\documentclass[final]{cvpr}

\usepackage{times}
\usepackage{epsfig}
\usepackage{graphicx}
\usepackage{amsmath}
\usepackage{amssymb}
\usepackage{adjustbox}
\usepackage{booktabs}
\usepackage{wrapfig}
\usepackage{multirow}
\usepackage{subcaption}
\usepackage[pagebackref=true,breaklinks=true,colorlinks,bookmarks=false]{hyperref}

\DeclareMathOperator*{\argmax}{arg\,max}

\newcommand{\myparagraph}[1]{\vspace{2pt}\noindent{\bf #1}}

\def\cvprPaperID{1098} % *** Enter the CVPR Paper ID here
\def\confYear{CVPR 2021}
\begin{document}

%%%%%%%%% TITLE
\title{Learning Graph Embeddings for Compositional Zero-shot Learning}

\author{\vspace{0em}
\setlength\tabcolsep{0.1em}
% \pagenumbering{gobble}
\begin{tabular}{cccc} 
Muhammad Ferjad Naeem$^{1,3}$, & Yongqin Xian$^2$, & Federico Tombari$^{3,4}$, & Zeynep Akata$^{1,2,5}$ \tabularnewline
\end{tabular}
\\
\renewcommand{\arraystretch}{0.5}
\begin{tabular}{ccccc} 
    $^1$\normalsize{University of Tübingen,} & $^2$\normalsize{MPI for Informatics,} & $^3$\normalsize{TUM,} & $^4$\normalsize{Google} & $^5$\normalsize{MPI for Intelligent Systems} %\tabularnewline
    %\normalsize{Yonsei University} & \normalsize{\enskip LINE Plus Corp.} & \normalsize{\enskip NAVER Corp.} & \normalsize{}
\end{tabular}
}%

\maketitle
\thispagestyle{empty}

%%%%%%%%% ABSTRACT
\begin{abstract}
In compositional zero-shot learning, the goal is to recognize unseen compositions~(e.g. old dog) of observed visual primitives states~(e.g. old, cute) and objects~(e.g. car, dog) in the training set. This is challenging because the same state can for example alter the visual appearance of a dog drastically differently from a car. 
As a solution, we propose a novel graph formulation called Compositional Graph Embedding (CGE) that learns image features, compositional classifiers and latent representations of visual primitives in an end-to-end manner. The key to our approach is exploiting the dependency between states, objects and their compositions within a graph structure to enforce the relevant knowledge transfer from seen to unseen compositions.
By learning a joint compatibility that encodes semantics between concepts, our model allows for generalization to unseen compositions without relying on an external knowledge base like WordNet. We show that in the challenging generalized compositional zero-shot setting our CGE significantly outperforms the state of the art on MIT-States and UT-Zappos. We also propose a new benchmark for this task based on the recent GQA dataset. Code is available at: \url{https://github.com/ExplainableML/czsl}
\end{abstract}

%%%%%%%%% BODY TEXT

\section{Introduction}
A ``black swan'' was ironically used as a metaphor in the 16th century for an unlikely event because the western world had only seen white swans. Yet when the European settlers observed a black swan for the first time in Australia in 1697, they immediately knew what it was. This is because humans posses the ability to compose their knowledge of known entities to generalize to novel concepts.
%This isn't by chance, 
Since visual concepts follow a long tailed distribution \cite{salakhutdinov2011learning, wang2017learning}, it is not possible to gather supervision for all concepts. Therefore, recognizing shared and discriminative properties of objects and reasoning about their various states has evolved as an essential part of human intelligence. 
Once familiar with the semantic meaning of these concepts, we can recognize unseen compositions of them without any supervision. While there is a certain degree of compositionality in modern vision systems, e.g. feature sharing, most models are not compositional in the classifier space and treat every class as an independent entity requiring training for any new concept.

\begin{figure}
    \centering
    \includegraphics[width= \columnwidth]{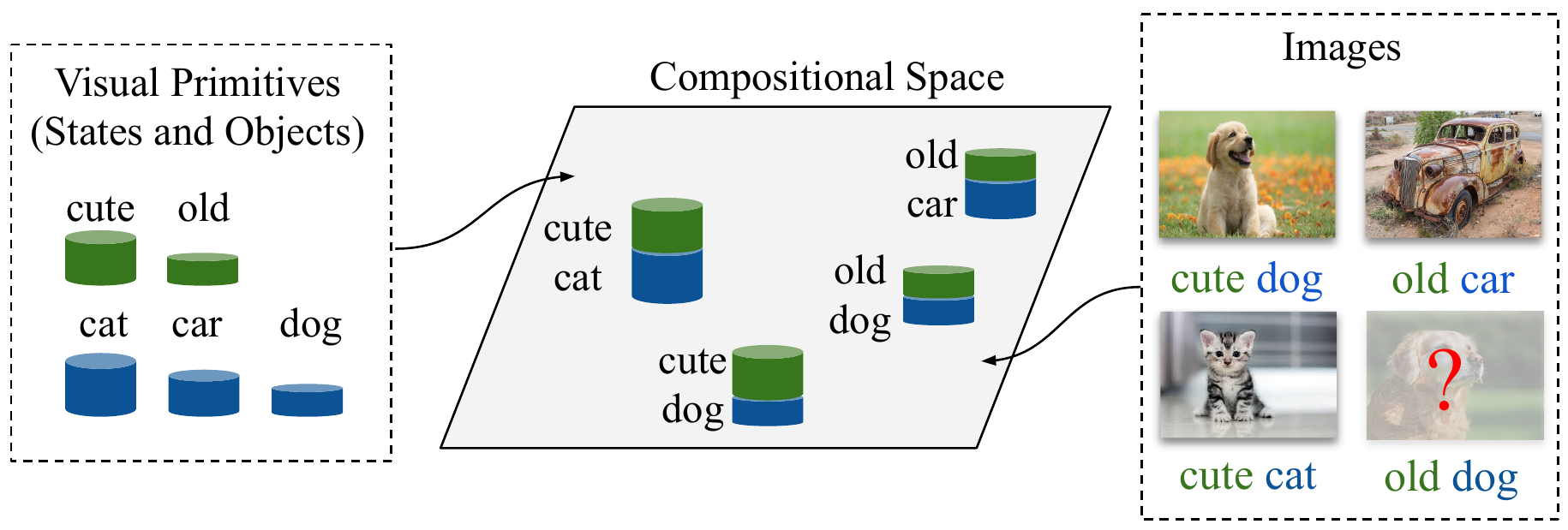}
    \caption{We aim to build a classifier for a novel state of a known object (e.g. \texttt{old dog}) given the knowledge of the shared primitives state and object in the training set.}
    \label{fig:teaser}
\end{figure}

In this work, we study the state-object compositionality problem also known as Compositional Zero-Shot Learning (CZSL)\cite{redwine}. The goal is to learn the compositionality of observed objects and their states as visual primitives to generalize to novel compositions of them as shown in figure~\ref{fig:teaser}. 
Some notable existing works in this field include learning a transformation network on top of individual classifiers~\cite{redwine}, treating states as linear transformations of object vectors~\cite{aopp}, learning modular networks conditioned on compositional classes~\cite{tmn} and learning object embeddings that are symmetric under different states~\cite{symnet}. 
However, these works treat each state-object composition independently, ignoring the rich dependency structure of different states, objects and their compositions. For example, learning the composition \texttt{old dog} is not only dependent on the state \texttt{old} and object \texttt{dog}, but also can be supported by other compositions like \texttt{cute dog}, \texttt{old car}, etc. We argue that such dependency structure provides a strong regularization which allows the network to better generalize to novel compositions. 
We therefore propose to exploit this dependency relationship by constructing a compositional graph to learn embeddings that are globally consistent.

Our contributions are as follows: (1) We introduce a novel graph formulation named Compositional Graph Embedding~(CGE) to model the dependency relationship of visual primitives and compositional classes. This graph can be created independently of an external knowledge base like WordNet~\cite{wordnet}. (2) Observing that visual primitives are dependent on each other and their compositional classes (figure~\ref{fig:teaser}), we propose a multimodal compatibility learning framework that learns to embed related states, objects and their compositions close to each other and far away from the unrelated ones.
(3) We propose a new benchmark called C-GQA for the task of CZSL. This dataset is curated from the recent GQA\cite{gqa} dataset with diverse compositional classes and clean annotations compared to datasets used in the community. 
(4) Our model significantly improves the state of the art on all the metrics on MIT-States, UT-Zappos and C-GQA datasets.

%-------------------------------------------------------------------------
\section{Related work}

\myparagraph{Compositionality} can loosely be defined as the ability to decompose an observation into its primitives. These primitives can then be used for complex reasoning. One of the earliest attempts in computer vision in this direction can be traced to Hoffman~\cite{hoffman1984parts} and  Biederman~\cite{biederman1987recognition} who theorized that visual systems can mimic compositionality by decomposing objects to their parts.
% .  later took a similar approach to define objects by their components. 
Compositionality at a fundamental level is already included in modern vision systems. 
Convolutional Neural Networks (CNN) have been shown to exploit compositionality by learning a hierarchy of features\cite{zeiler2014visualizing, lecun1989backpropagation}. Transfer learning\cite{caruana1997multitask, choi2013adding, deng2014large, patricia2014learning}  and few-shot learning\cite{hariharan2017low, ravi2016optimization, mensink2012metric} exploit the compositionality of pretrained features to generalize to data constraint environments. Visual scene understanding\cite{johnson2015image, dai2017detecting, jae2018tensorize, lu2016visual} aims to understand the compositionality of concepts in a scene.
Nevertheless, these approaches still requires collecting data for new classes.

\myparagraph{Zero-Shot Learning} aims at recognizing novel classes that are not observed during training~\cite{LNH13}. This is accomplished by using side information that describes novel classes e.g. attributes~\cite{LNH13}, text descriptions~\cite{RALS16} or word embeddings~\cite{SGMN13}. Some notable approaches include learning a compatibility function between image and class embeddings~\cite{akata2013label, zhang2016learning} and learning to generate image features for novel classes~\cite{xian2018feature, Zhu_2018_CVPR}. Graph convolutional networks~(GCN)~\cite{gcn, gcnzs, gcnzsrethinking} have shown to be promising for zero-shot learning. Wang et al.~\cite{gcnzs} propose to directly regress the classifier weights of novel classes with a GCN operated on an external knowledge graph~(WordNet~\cite{wordnet}). Kampffmeyer et al.\cite{gcnzsrethinking} improve this formulation by introducing a dense graph to learn a shallow GCN as a remedy for the laplacian smoothing problem~\cite{li2018deeper}. 

\myparagraph{Graph Convolutional Networks} are a special type of neural networks that exploit the dependency structure of data~(nodes) defined in a graph. 
Current methods~\cite{gcn} are limited by the network depth due to over smoothing at deeper layers of the network. The extreme case of this can cause all nodes to converge to the same value~\cite{li2018deeper}. Several works have tried to remedy this by dense skip connections~\cite{xu2018representation, li2019deepgcns}, randomly dropping edges~\cite{rong2019dropedge} and applying a linear combination of neighbor features~\cite{wu2019simplifying, klicpera2019diffusion, klicpera2018predict}. A recent work in this direction from Chen et al.\cite{gcnii} combines residual connections with identity mapping.

\myparagraph{Compositional zero-shot learning} stands at the intersection of compositionality and zero-shot learning and focuses on state and object relations.
We aim to learn the compositionality of objects and their states from the training set and are tasked with generalizing to unseen combination of these primitives.
Approaches in this direction can be divided into two groups. The first group is directly inspired by \cite{hoffman1984parts, biederman1987recognition}. Some notable methods including learning a transformation upon individual classifiers of states and objects~\cite{redwine}, modeling each state as a linear transformation of objects~\cite{aopp}, learning a hierarchical decomposition and composition of visual primitives\cite{yang2020learning} and modeling objects to be symmetric under attribute transformations\cite{symnet}.
An alternate line of works argues that compositionality requires learning a joint compatibility function with respect to the image, the state and the object\cite{causal, tmn, wang2019task}. This is achieved by learning a modular networks conditioned on each composition \cite{tmn,wang2019task} that can be ``rewired" for a new compositions. Finally a recent work from Atzmon et al. \cite{causal} argue that achieving generalization in CZSL requires learning the causality of visual transformation through a causal graph where the latent representation of primitives are independent of each other. 

%------------------------------------------------------------------------
\begin{figure*}
    \centering
    \includegraphics[width = \linewidth]{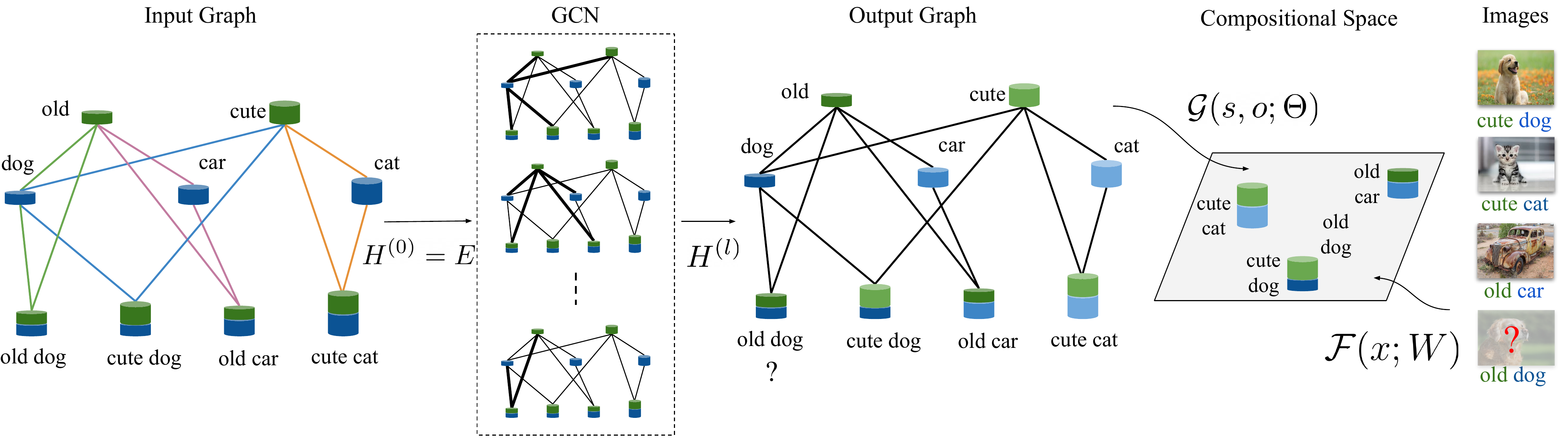}
    \caption{Compositional Graph Embed (CGE) learns a globally consistent joint embedding space between image features and classes of seen and unseen compositions from a graph.  
    In our novel graph formulation, nodes are connected if a dependency exists in form of a compositional label e.g. old, car and old car.
    We backpropagate the classification loss through the seen compositional nodes to the GCN $\mathcal{G}$ and the feature extractor $\mathcal{F}$.
    Hence, the representation of e.g. the dog is compatible with its different states and the representation of old dog aggregates the knowledge from old, dog, cute dog, old car etc. }
    \label{fig:method}
\end{figure*}

Our proposed method lies at the intersection of several discussed approaches. We learn a joint compatibility function similar to \cite{causal, tmn, wang2019task} and utilize a GCN similar to \cite{gcnzs, gcnzsrethinking}. However, our approach exploits the dependency structure between states, objects and compositons which has been overlooked by previous CZSL approaches~\cite{causal, tmn, wang2019task}. 
Instead of using a predefined knowledge graph like WordNet~\cite{wordnet} to regress pretrained classifiers of the seen classes~\cite{gcnzs, gcnzsrethinking}, we propose a novel way to build a compositional graph and learn classifiers for all classes in an end-to-end manner. In contrast to Atzmon et al.\cite{causal} we explicitly promote the dependency between all primitives and their compositions in our graph. This allows us to learn embeddings that are consistent with the whole graph. Finally, unlike all existing methods~\cite{redwine, aopp, causal, tmn, wang2019task, yang2020learning}, we do not rely on a fixed image feature extractor and train our pipeline end-to-end.

\section{Approach}
We consider the image classification task where each image is associated with a label that is composed of a state~(e.g. \texttt{cute}) and an object~(e.g. \texttt{dog}). The goal of compositional zero-shot learning~(CZSL)~\cite{redwine} is to recognize the compositional labels that are not observed during training. This is particularly challenging as
the states significantly change the visual appearance of an object hindering the performance of the classifiers.

We propose a novel formulation to the problem, namely Compositional Graph Embedding~(CGE), which constructs a compositional graph and adopts a graph convolutional network to learn the dependency structure between labels. 
An overview of our approach is shown in Figure~\ref{fig:method}. It builds on the compatibility learning framework that learns a class-agnostic scoring function between an image and a compositional label. The input image is encoded with an image feature extractor $\mathcal{F}$, while the classifier weights for the compositional label are learned by a composition function $\mathcal{G}$. The key insight of our approach is that leveraging the dependency relationship between states, objects and their compositions is beneficial for recognizing unseen compositions.   

\subsection{Compatibility Learning Framework for CZSL}

\myparagraph{Task formulation.} We formalize the CZSL task as follows. Let $\mathcal{T}=\{(x, y) | x\in \mathcal{X}, y\in\mathcal{Y}_s \}$ where $\mathcal{T}$ stands for the training set, $x$ denotes an image in the RGB image space $\mathcal{X}$ and $y$ is its label belonging to one of the seen labels $\mathcal{Y}_s$. Each label is a tuple $y=(s,o)$ of a state $s\in \mathcal{S}$ and an object $o\in \mathcal{O}$ with $\mathcal{S}$ and $\mathcal{O}$ being the set of states and objects respectively. The task of CZSL is to predict a set of novel labels $\mathcal{Y}_n$ that consists of novel compositions of states $\mathcal{S}$ and objects $\mathcal{O}$ i.e., $\mathcal{Y}_s\cap \mathcal{Y}_n=\emptyset$. Following \cite{tmn, zslxian18benchmark}, we study this problem in the generalized compositional zero-shot setting where the test set includes images from both seen and novel compositional labels %and the model has to predict over the combinations of seen and unseen compositional classes
$\mathcal{Y}=\mathcal{Y}_s\cup\mathcal{Y}_n$.

\myparagraph{Compatibility function.} Learning state and object classifiers separately is prone to overfit to labels observed during training because states and objects are not independent e.g. the appearance of the state \texttt{sliced} varies significantly with the object (e.g. \texttt{apple} or \texttt{bread}). Therefore, we chose to model them jointly by learning a compatibility function $f: \mathcal{X} \times \mathcal{S}\times \mathcal{O} \longrightarrow \mathbb{R}$ that captures the compatibility score between an image, a state and an object. Given a specific input image $x$, we predict its label $y=(s, o)$ by searching the state and object composition that yields the highest compatibility score:
\begin{equation}
    \label{eq:compatibility}
    f(x, s, o) = \mathcal{F}(x; W)\cdot \mathcal{G}(s, o; \Theta)
\end{equation}
where $\mathcal{F}(x; W)\in \mathbb{R}^{d}$ is the image feature extracted from a pretrained feature extractor, $\mathcal{G}(s, o; \Theta)\in \mathbb{R}^{d}$ is a function that outputs the label embedding of the state-object pair $(s, o)$,  ($W$,$\Theta$) are respectively the learnable parameters of $\mathcal{F}$ and $\mathcal{G}$, and ($\cdot$) is the dot product. The compatibility function assigns high scores to the correct triplets i.e., image $x$ and its label $(s, o)$, and low scores to the incorrect ones. The label embedding can be also interpreted as the classifier weights for the label $(s, o)$ and we use the two terms interchangeably. 

Our compatibility learning framework is closely related to \cite{redwine, tmn}. LabelEmbed~\cite{redwine} parameterizes the compositional embedding function with a multi-layer perceptron and computes the compositions from the word embeddings~(e.g. word2vec~\cite{word2vec}) of states and objects, while TMN~\cite{tmn} adopts a modular network as the image feature extractor and a gating network as the compositional embedding function. We argue that there exists a complex dependency structure between states, objects and their compositions and learning this dependency structure is crucial. To this end, we propose to integrate the compositional embedding function $\mathcal{G}$ as a graph convolutional neural network~(GCN) which adds an inductive bias to the inherent structure between states, objects, and their combination defined by our compositional graph introduced next. 

\subsection{Compositional Graph Embedding~(CGE)}
We propose the Compositional Graph Embedding~(CGE) framework integrating the Graph Convolutional Networks~(GCN)~\cite{gcn} to the compositional embedding function $G(s, o)$ that learns the label embedding for each compositional label $y=(s, o)\in \mathcal{Y}$ in an end to end manner. The GCN network exploits the dependency structure in a predefined compositional graph from states, objects and their compositions~(including both seen and unseen labels). In the following, we first define the compositional graph, then introduce the node features and finally explain how to learn a GCN for the CZSL task.

\myparagraph{Compositional graph.} 
Our graph consists of $K=|\mathcal{S}|+|\mathcal{O}|+|\mathcal{Y}|$ nodes that represent states $\mathcal{S}$,  objects $\mathcal{O}$ and compositional labels $\mathcal{Y}$. Two nodes are connected if they are related. The key insight of our graph is that each compositional label $y=(s, o)\in \mathcal{Y}$ defines a dependency relationship between the state $s$, object $o$ and their composition $y$. To this end, we build the edges of the graph by connecting $(s, o)$, $(s, y)$ and $(o, y)$ for every $y=(s, o)\in \mathcal{Y}$. In addition, each node is also connected to itself. Note that the edges in our graph are unweighted and undirected, leading to a symmetric adjacency matrix $L\in \mathbb{R}^{K\times K}$ where element $L_{ij}=1$ if there is a connection between nodes $i$ and $j$ otherwise $L_{ij}=0$. Despite its simplicity, we find that our compositional graph provides the accurate dependency structure to recognize unseen compositional labels. 

\myparagraph{Node features.} GCN~\cite{gcn,gcnii} operates on node features in a neighborhood defined by the graph. Therefore, after obtaining the compositional graph, we need to represent each node with a proper feature embedding. We chose to use the word embeddings~\cite{word2vec,fasttext} pretrained on a large text corpus e.g. Wikipedia, because they capture rich semantic similarities among words i.e., \texttt{dog} is closer to \texttt{cat} than to \texttt{car} in the word embedding space. Specifically, every state or object node in the compositional graph is represented by the word embedding associated to its corresponding state or object name. We compute the node features of the the compositional label~(e.g. \texttt{cute dog}) by averaging the word embeddings of the corresponding state~(e.g. \texttt{cute}) and object~(e.g. \texttt{dog}) names. As indicated in~\cite{word2vec}, by adding word embeddings we achieve compositionality in the semantic space. We represent the input node features with a matrix $E\in \mathbb{R}^{K\times P}$ where $K$ is the total number of nodes and each row denotes the $P$-dim feature of a graph node. 

\myparagraph{Graph convolutional network for CZSL.}
 GCN~\cite{gcn} is an efficient multi-layer network to learn new feature representation of nodes for a downstream task that are consistent with the graph structure. Here, we apply the GCN to tackle the CZSL task by directly predicting the compositional label embeddings. The input of our GCN consists of the compositional graph, represented by the adjencency matrix $L\in \mathbb{R}^{K\times K}$ and the node feature matrix $E\in \mathbb{R}^{K\times P}$. Specifically, each GCN layer computes the hidden representation of each node by convolving over neighbor nodes using a simple propagation rule~\cite{gcn} also known as a spectral convolution,
\begin{equation}
    H^{(l+1)} = \sigma (D^{-1} L H^{(l)} \Theta^{(l)})
    \label{eq:gcn}
\end{equation}
where $\sigma$ represents the non-linearity activation function ReLU, $H^{(l)}\in \mathbb{R}^{K\times U}$ denotes the hidden representations in the $l^{th}$ layer with $H^{(0)}=E$ and $\Theta\in \mathbb{R}^{U\times V}$ is the trainable weight matrix with $V$ learnable filters operating over $U$ features of $H^{(l)}$. $D \in \mathbb{R}^{K\times K}$ is a diagonal node degree matrix which normalizes rows in $L$ to preserve the scale of the feature vectors. 
By stacking multiple such layers, the GCN propagates the information through the graph to obtain better node embeddings for both the seen and unseen compositional labels. For example, our GCN allows an unseen compositional label e.g. \texttt{old dog} to aggregate information from its neighbor nodes e.g.  \texttt{old}, \texttt{dog}, \texttt{cute dog}, and \texttt{old car} that are observed~(see Figure~\ref{fig:method}). 

\myparagraph{Objective.} As the objective of the GCN is to predict the classifier weights of the compositional labels, the node embedding of the output layer in the GCN has the same dimentionality as the image feature $\mathcal{F}(x)$. This indicates that our compositional embedding function becomes $\mathcal{G}(s, o)=H_y^{(N)}$ where $H^{(N)}$ is the output node embedding matrix and $H_y^{(N)}$ denotes the row corresponding to the compositional label $y=(s, o)$. We then optimize the following cross-entropy loss to jointly learn the image feature extractor and GCN in an end-to-end manner,
\begin{equation}
\label{eq:celoss}
    \min_{W, \Theta} \frac{1}{|\mathcal{T}|} \sum_{i=1}^{|\mathcal{T}|} -log ( \frac{\exp{f(x_i, s_i, o_i)}}{\sum_{j\in\mathcal{Y}_s} \exp{f(x_i, s_j, o_j)}}  )
\end{equation}
where $f$ is the compatibility function defined in Equation~\ref{eq:compatibility}, $y=(s_i, o_i)$ is the ground truth label of image $x_i$, label  $y^\prime=(s_j, o_j)$ denotes any seen compositional class, $W$ and $\Theta$ are the learnable parameters of the feature extractor and the GCN respectively. Intuitively, the cross-entropy loss enables the compatibility function to assign the high scores for correct input triplets. 

\myparagraph{Inference.} At test time, given an input image $x$, we derive a prediction by searching the compositional label that yields the highest compatibility score, 
\begin{equation}
    \argmax_{y=(s, o) \in \mathcal{Y}} f(x, s, o).
\end{equation}
It is worth noting that our model works in the challenging generalized CZSL setting~\cite{tmn}, where both seen and unseen compositional classes~(i.e. $\mathcal{Y}=\mathcal{Y}_s\cup\mathcal{Y}_n$) are predicted.

\myparagraph{Discussion.} To the best of our knowledge, our Compositional Graph Embedding~(CGE) is the first end-to-end learning method that jointly optimizes the feature extractor $\mathcal{F}$ and the compositional embedding function $\mathcal{G}$ for the task of compositional zero-shot learning. 

Compared to prior CZSL works~\cite{tmn,symnet,redwine,aopp} our CGE does not overfit while optimizing the CNN backbone of $\mathcal{F}$ (see supplementary) as it is regularized by the compositional graph that defines the dependency relationship between classes making the end-to-end training beneficial. 
Compared to previous GCN work~\cite{gcnzs, gcnzsrethinking} that utilizes GCNs to regress the fixed classifier weights to learn classifiers of novel classes, we directly use image information to learn classifiers for both seen and novel classes. Moreover, while \cite{gcnzs, gcnzsrethinking} rely on a known knowledge graph like WordNet\cite{wordnet} describing the relation of novel classes to seen classes, our CGE cannot rely on existing knowledge graphs like WordNet\cite{wordnet} because they do not cover compositional labels. Therefore, we propose to construct the graph by exploiting the dependency relationship defined in the compositional classes. 
We find that propagating information from seen to unseen labels through this graph is crucial for boosting the CZSL performance.

%-------------------------------------------------------------------------

{
\setlength{\tabcolsep}{2pt}
\renewcommand{\arraystretch}{1.4}
\begin{table}[t]
    \centering
    \resizebox{\linewidth}{!}
    {\begin{tabular}{l cc|cc|ccc|ccc}
        %\toprule
         &   &  & \multicolumn{2}{c}{\textbf{Training}} & \multicolumn{3}{c}{\textbf{Validation}} & \multicolumn{3}{c}{\textbf{Test}}\\
        %\toprule
        \textbf{Dataset}& s & o & sp  & i & sp  & up  & i & sp  & up  & i \\ 
    \hline
    MIT-States\cite{mitstates}     &  115 & 245 & 1262 & 30k & 300 & 300 & 10k & 400 & 400 & 13k \\
    UT-Zappos\cite{utzappos1} & 16 & 12 & 83 & 23k & 15 & 15 & 3k & 18 & 18 & 3k \\
    C-GQA (Ours) & 453 & 870 & 6963 & 26k & 1173 & 1368 & 7k & 1022 & 1047 & 5k \\
    %\bottomrule
    \end{tabular}}
    \caption{\textbf{Dataset statistics for CZSL}: We use three datasets to benchmark our method against the baselines. C-GQA (ours): our proposed dataset splits from Stanford GQA dataset \cite{gqa}. (s: \# states, o: \# objects, sp: \# seen compositions, up: \# unseen compositions, i: \# images)}
    \label{tab:dataset}
\end{table}
}

{
\setlength{\tabcolsep}{2pt}
\renewcommand{\arraystretch}{1.3}
\begin{table*}[t]
    \centering
    \resizebox{\linewidth}{!}
    {
    \begin{tabular}{l|ccccccc|ccccccc|ccccccc}
    %\toprule
    & \multicolumn{7}{c}{\textbf{MIT-States}} & \multicolumn{7}{c}{\textbf{UT-Zap50K}} &
    \multicolumn{7}{c}{\textbf{C-GQA}}\\
    %\toprule
    \multirow{2}{*}{\textbf{Method}} & \multicolumn{2}{c}{AUC}  &  \multicolumn{3}{c}{Best} & & &
    \multicolumn{2}{c}{AUC}  &   \multicolumn{3}{c}{Best} & & &
    \multicolumn{2}{c}{AUC}    & \multicolumn{3}{c}{Best}  &  & \\
    & Val& Test & HM &  Seen & Unseen  & s & o & 
    Val& Test & HM &  Seen & Unseen  & s & o & 
    Val& Test & HM &  Seen & Unseen  & s & o \\
    \hline
    AttOp\cite{aopp} &  2.5 & 1.6 & 9.9 & 14.3 & 17.4  & 21.1 & 23.6 &
    21.5 & 25.9 & 40.8 & 59.8 & 54.2&  38.9 & 69.6 &  
     0.9 & 0.3  & 2.9 & 11.8 & 3.9 & 8.3 & 12.5\\
    LE+\cite{redwine} & 3.0 & 2.0 & 10.7 & 15.0 & 20.1 &  23.5 & 26.3 & 
    26.4 & 25.7 & 41.0 & 53.0 & 61.9& 41.2 & 69.2 &   
    1.2 &  0.6 & 5.3 & 16.1  & 5.0 & 7.4 & 15.6  \\
    TMN\cite{tmn} & 3.5 & 2.9 & 13.0 & 20.2 & 20.1 &  23.3 & 26.5 & 
    36.8 & 29.3 & 45.0 & 58.7 & 60.0 & 40.8 & 69.9 &   
     2.2&  1.1 & 7.7 & 21.6 & 6.3 & 9.7  &20.5\\
    SymNet\cite{symnet} & 4.3 & 3.0 & 16.1 & 24.4 & 25.2 &  26.3 & 28.3 & 
    25.9 & 23.9 & 39.2 & 53.3 & 57.9 &  40.5 & 71.2 &
    3.3 & 1.8  & 9.8 & 25.2 & 9.2 &14.5  & 20.2\\
    CGE\textsubscript{ff} (ours) &  6.8 & 5.1 & 17.2 & 28.7 & 25.3 &   27.9 & 32.0 & 
    38.7& 26.4 & 41.2 & 56.8 & 63.6 &   45.0 & 73.9 & 
    3.6 &  2.5 & 11.9 & 27.5 & 11.7 & 12.7 &26.9\\\hline
    \textbf{CGE (ours)} &  \textbf{8.6} & \textbf{6.5}& \textbf{21.4} & \textbf{32.8} & \textbf{28.0} &  \textbf{30.1} & \textbf{34.7} &  
    \textbf{43.2} & \textbf{33.5}& \textbf{60.5}  & \textbf{64.5} & \textbf{71.5} &  \textbf{48.7} & \textbf{76.2} &
    \textbf{5.0}&  \textbf{3.6} & \textbf{14.5} & \textbf{31.4} & \textbf{14.0} & \textbf{15.2}  & \textbf{30.4} \\
    \end{tabular}}
    \caption{\textbf{Comparison with the state of the art:} We compare our Compositional Graph Embed (CGE) with the state of the art on Validation and Test AUC~(in $\%$); best unseen, seen and harmomic mean (HM) accuracies~(in $\%$) as well as state (s) and object (o) prediction accuracies~(in $\%$) on widely used MIT-States and UT-Zappos datasets as well as our proposed C-GQA dataset.}
    \label{tab:basenumbers}
\end{table*}
}

\section{Experiments}
\label{sec:experiment}
After introducing our experimental setup, we compare our results with the state of the art, ablate over our design choices and present some qualitative results.

\myparagraph{Datasets.}
We perform our experiments on three datasets (see detailed statistics in Table \ref{tab:dataset}).
MIT-States\cite{mitstates} consists of natural objects in different states collected using an older search engine with limited human annotation leading to significant label noise \cite{causal}. UT-Zappos\cite{utzappos1, utzappos2} consists of images of a shoes catalogue which is arguably not entirely compositional as states like \textit{Faux leather} vs \textit{Leather} are material differences not always observable as visual transformations. We use the GCZSL splits from \cite{tmn}. 

To address the limitations of these two datasets,we propose a split built on top of Stanford GQA dataset \cite{gqa} originally proposed for VQA and name it Compositional GQA (C-GQA) dataset (see supplementary for the details). C-GQA contains over 9.5k compositional labels making it the most extensive dataset for CZSL. With cleaner labels and a larger label space, our hope is that this dataset will inspire further research on the topic. Figure \ref{fig:qual} shows some samples from the three datasets.

\myparagraph{Metrics.} 
As the models in zero-shot learning problems are trained only on seen $\mathcal{Y}_s$ labels (compositions), there is an inherent bias against the unseen $\mathcal{Y}_n$ labels. As pointed out by \cite{chao2016empirical, tmn}, the model thus needs to be calibrated by adding a scalar bias to the activations of the novel compositions to find the best operating point and evaluate the generalized CZSL performance \cite{tmn} for a more realistic setting.

We adopt the evaluation protocol of \cite{tmn} and report the Area Under the Curve (AUC)~(in $\%$) between the accuracy on seen and unseen compositions at different operating points with respect to the bias. The best unseen accuracy is calculated when the bias term is large leading to predicting only the unseen labels, also known as zero-shot performance. In addition, the best seen (base class) performance is calculated when the bias term is negative leading to predicting only the seen labels. As a balance between the two, we also report the best harmonic mean. 
To emphasize that this is different from the traditional zero-shot learning evaluation, we add the term ``best'' in our results. Finally, we report the state and object accuracy on the novel labels to show the improvement in classifying the visual primitives. 
We emphasize that the dataset splits we propose for C-GQA and use from \cite{tmn} for MIT-States and UT-Zappos do no not violate the zero-shot assumption as results are ablated on the validation set. Some works in CZSL use older splits that lack a validation set and thus use indirect full label supervision\cite{zslxian18benchmark} by ablating over the test set. We therefore advice future works to rely on the new splits.

\myparagraph{Training details.}
To be consistent with the state of the art, we use a ResNet18~\cite{resnet} backbone pretrained on ImageNet as the image feature extractor $\mathcal{F}$.  For a fair comparison with the models that use a fixed feature extractor, we introduce a simplification of our method named $CGE_{ff}$. We learn a 3 layer fully-connected (FC) network with ReLU\cite{nair2010rectified}, LayerNorm\cite{ba2016layer} and Dropout\cite{srivastava2014dropout} while keeping the feature extractor fixed for this baseline. 
We use a shallow 2-layer GCN with a hidden dimension of $4096$ as $\mathcal{G}$ (detailed ablation on this is presented in section \ref{sec:graphablataion}). On MIT-States, we initialize our word embeddings with a concatenation of pretrained fasttext\cite{fasttext} and word2vec models\cite{word2vec} similar to \cite{xian2019semantic}. On UT-Zappos and C-GQA, we initialize the word embeddings with word2vec(ablation reported in supplementary).

We use Adam Optimizer\cite{kingma2014adam} with a learning rate of $5e^{-6}$ for $\mathcal{F}$ and a learning rate of $5e^{-5}$ for $\mathcal{G}$. We implement our method in PyTorch\cite{paszke2019pytorch} and train on a Nvidia V100 GPU. For state-of-the-art comparisons, we use the authors' implementations where available. The code for our method and the new dataset C-GQA will be released upon acceptance.

%%%%%%%%%%%%%%%%%%%%%%%%%%%%%%%
\subsection{Comparing with the State of the Art}

We compare our results with the state of the art in Table \ref{tab:basenumbers} and show that our Compositional Graph Embed(CGE) outperforms all previous methods by a large margin and establishes a new state of the art for Compositional Zero-shot Learning. Our detailed observations are as follows.

\myparagraph{Generalized CZSL performance.}
Our framework demonstrates robustness against the label noise on MIT-States noted previously in \cite{causal}. For the generalized CZSL task, our CGE achieves a test AUC of 6.5\% which is an improvement of over 2$\times$ compared to the last best 3.0\% from SymNet. Similarly, as our method does not only improve results on seen labels but also unseen labels, it significantly boosts the state of the art harmonic mean, i.e. 16.1\% to 21.4\%. When it comes to state and object prediction accuracy, we observe an improvement from 26.3\% to 30.1\% for states and 28.3\% to 34.7\% for objects. Although our results significantly improve the state of the art on all metrics, the state and object accuracies are quite low, partially due to the label noise for this dataset. 

Similar observations are confirmed on UT-Zappos, where we achieve a significant improvement on the state of the art with an AUC of 33.5\% compared to 29.3\% from TMN.
An interesting observation is that SymNet, i.e. the state of the art on MIT States, with an AUC of 23.9\% does not achieve the best performance in the generalized CZSL setting on UT Zappos. We conjecture that this is because the state labels in this dataset are not entirely representing visual transformations, something this method was designed to exploit. In this dataset, our fully compositional model improves the best harmonic mean wrt the state of the art significantly (45.0\% with TMN vs 60.5\% ours). Note that, this is due to a significant accuracy boost achieved on unseen compositions (60.0\% vs 71.5\%). 

Finally on the proposed splits of the GQA dataset~\cite{gqa}, i.e. C-GQA dataset, we achieve a test AUC of 3.6\% outperforming the closest baseline by a 2$\times$. Note that, since C-GQA has a compositional space of over 9.3k concepts, it is significantly harder than MIT-States and UT-Zappos while being truly compositional and containing cleaner labels. The state and the object accuracies of our method are 15.2\% and 30.4\%, \ie significantly higher than the state of the art. However these results also indicate the general difficulty of the task. Similarly, our best seen and best unseen accuracies (31.4\% and 14.0\%) indicate a large room for improvement on this dataset, which may encourage further research with our C-GQA dataset on the CZSL task.

We also make an interesting observation on all three datasets. While SymNet uses an object classifier that is trained independently from the compositional pipeline, our method consistently outperforms it on object accuracy.
We conjecture that this is because a compositional network sensitive to the information about the states is also a better object classifier, since it disentangles what it means to be an object from the state it is observed in, preventing biases to properties like textures~\cite{biasimagenet}. This insight can be an avenue for future improvement in object classification.

\myparagraph{Impact of feature representations.} 
To quantify the improvement of our graph formulation on the same feature representations as the state of the art, we also present results of our CGE with a fixed feature extractor ~(Resnet18), i.e. denoted by $CGE_{ff}$, in Table \ref{tab:basenumbers}. 
We see that this version of our model also consistently outperforms the state of the art by a large margin on MIT-States and C-GQA while matching the performance on UT-Zappos. Especially on MIT-States, the improvement over the state of the art is remarkable, i.e. 5.1\% test AUC vs 3.0\% test AUC with SymNet.
In summary, this shows that our method benefits from both the knowledge propagation in the compositional graph and from learning better image representations.

For a fair comparison, we also allowed the previous baselines to train end-to-end with $\mathcal{F}$. However, this results in a significant performance drop indicating they are unable to jointly learn the feature extractor against the RGB space. To address this limitation, some works\cite{xian2018feature, wang2019task} have proposed to use a generative network to learn the distribution of image features in zero-shot problems. We, on the other hand, don't need to rely on an expensive generative network and jointly learn the image representations and the compositional classifiers in an end-to-end manner.

\subsection{Ablation study}
\label{sec:ablation}
In this section we ablate our CGE model with respect to the graph connections, the graph depth and graph convolution variants. 
%%%%%%%%%%%%%%%%%%%%%%%%
{
\setlength{\tabcolsep}{6pt}
\renewcommand{\arraystretch}{1.2}
\begin{table}[t]
    \centering
    \resizebox{\linewidth}{!}
    {\begin{tabular}{l|cc}
%\toprule
\textbf{Connections in Graph} & \textbf{AUC} & \textbf{Best HM}   \\
\hline
%\cmidrule[1pt]{1-3}
a) Direct Word Embedding & 5.9 & 19.4 \\
b) (s,y) and (o,y), no self-loop on y  & 7.6 & 18.6 \\
c) (s,y) and (o,y) & 8.1 & 22.7 \\
d) \textbf{\textit{CGE}:} (s,y), (o,y) and (s,o)    & \textbf{8.6} & \textbf{23.3} \\
e) Extra WordNet hierarchy on $\mathcal{O}$ & 7.9 & 22.0 \\
%\bottomrule
\end{tabular}}
\caption{\textbf{Ablation over the graph connections} validates the structure of our proposed graph on the validation set of MIT-States dataset. We start from directly using the word embeddings as classifier weights to learning a globally consistent embedding from a GCN as the classifier weights (s: states, o: objects, y: compositional labels). }
% \ferjad{add connections}
\label{tab:gcnconnection}
\end{table}
}

\myparagraph{Graph connections.} 
We perform an ablation study with respect to the various connections in our compositional graph on the validation set of MIT-States and report results in Table \ref{tab:gcnconnection}. In the Direct Word Embedding variant, i.e. row (a) our label embedding function $\mathcal{G}$ is an average operation of state and object word embeddings. We see that, directly using word embedding of compositional labels as the classifier weights leads to an AUC of 5.9. 
In row (b) we represent a graph with connections between states (s) to compositional labels (y) and objects to compositional labels (y) but remove the self connection for the compositional label. In this case, the final representation of compositional labels from the GCN only combines the hidden representations of states and objects leading to an AUC of 7.6. 

Row (c) represents the graph that has self connections from each compositional label in addition to the connections between states and compositional labels as well as objects and compositional labels as in row (b). We see that this variant achieves an AUC of 8.1 indicating that the hidden representation of compositional classes is beneficial.

Row (d) is our final model where we additionally incorporate the connections between states and objects in a pair to model the dependency between them. We observe that learning a representation that is consistent with states, objects and the compositional labels increases the AUC from 8.1 to 8.6 validating our choice of connections in the graph. We again emphasize that in the absence of an existing knowledge graph for compositional relations, our simple but well designed graph structure is able to capture the dependencies between various concepts. 

While our final CGE does not employ an external knowledge graph, we can utilize an existing graph like WordNet~\cite{wordnet} to get the hierarchy of the object classes similar to some baselines in zero-shot learning~\cite{gcnzs,gcnzsrethinking}. Row (e) represents a model that exploits object hierarchy in addition to our compositional graph discussed earlier. This leads to additional 418 nodes to model the parent child relation of the objects. 
% Although interesting, 
We see that this results in a slight performance drop with an AUC of 7.9 because this graph may not be compatible with the compositional relations.

%%%%%%%%%%%%%%%%%%%%%%%
\begin{figure}[t]
    \centering
    \includegraphics[width = \columnwidth]{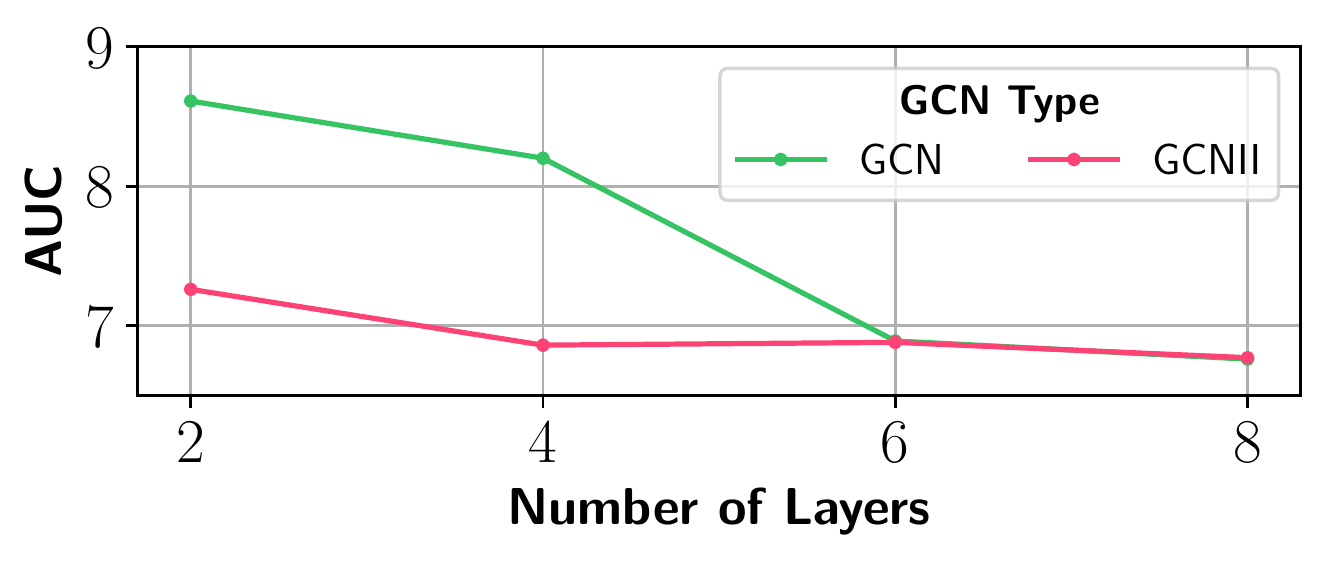}
    \caption{
    \textbf{Graph convolution and depth:} We compare the spectral convolution GCN\cite{gcn} with the recent GCNII\cite{gcnii} that aims to address the over smoothing issue at increasing depth. We perform the comparison at various depths of the GCN network on the validation set of MIT-States.}
    \label{fig:gcnvsgcnii}
\end{figure}
%%%%%%%%%%%%%%%%%%%%%

%%%%%%%%%%%%%%%%%%%%%
\begin{figure*}[t]
    \begin{subfigure}{0.67\textwidth}
         \centering
    \includegraphics[width=\textwidth]{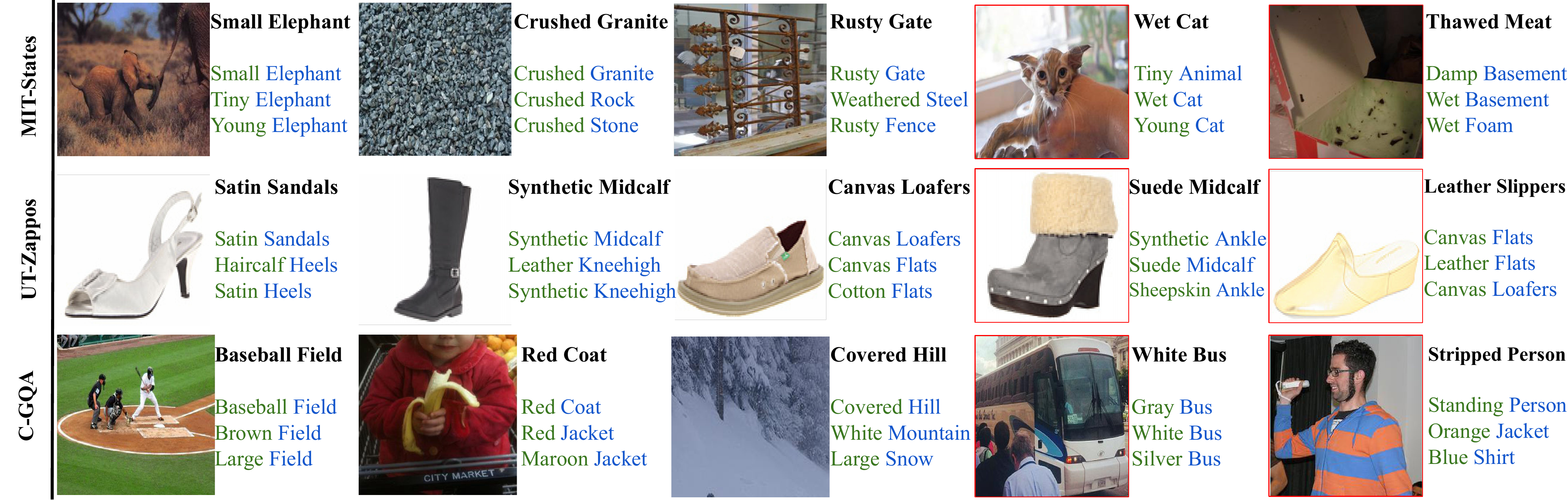}
    % \vspace{2mm}
    \caption{Retrieving compositional labels}
    \end{subfigure}
    %\hspace{2mm}
    \begin{subfigure}{0.32\textwidth}
         \centering
    \includegraphics[width= \columnwidth]{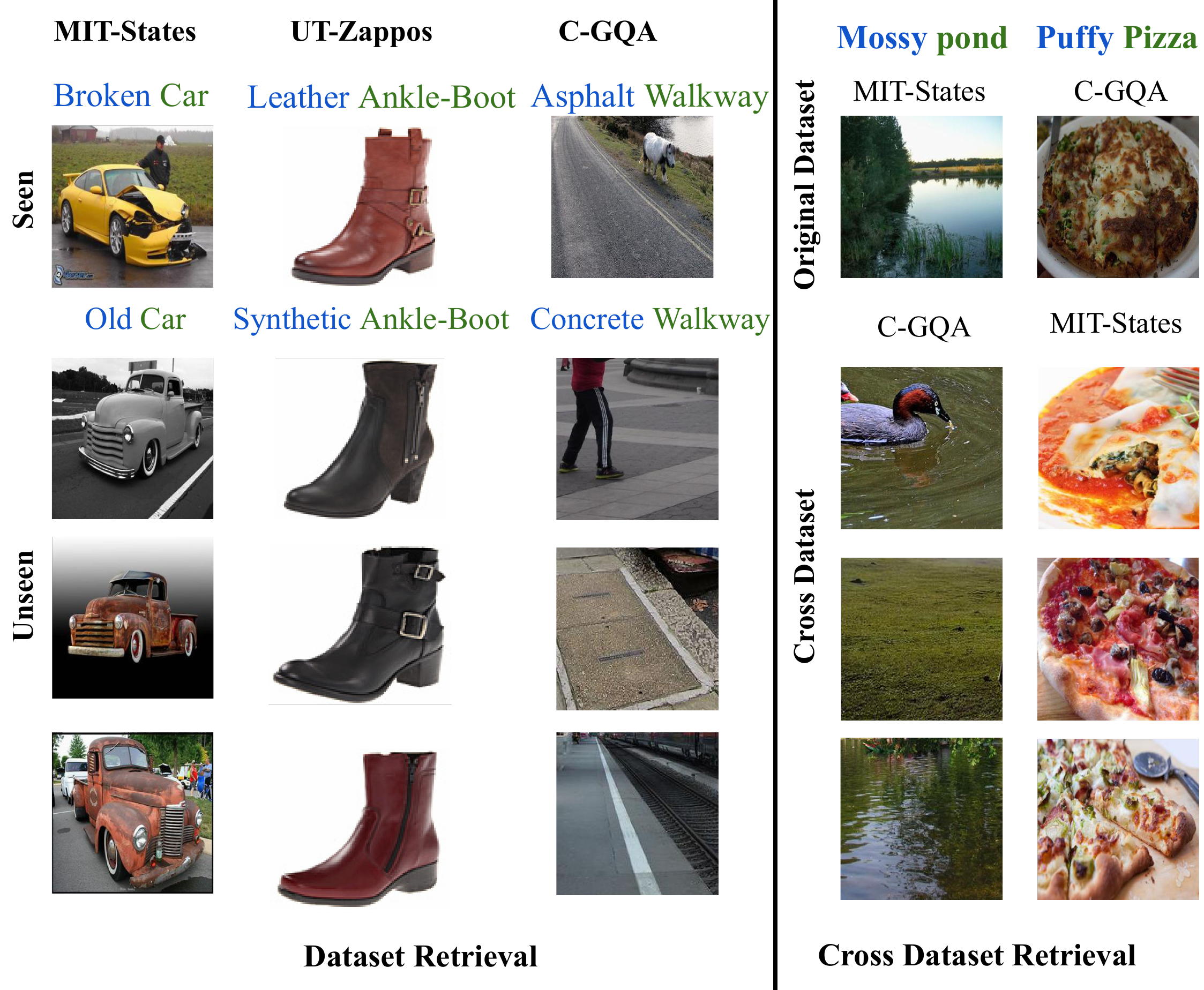}
    \caption{Retrieving images}
    \end{subfigure}
    \caption{\textbf{Qualitative results.} Left: We show the top-3 predictions of our model for some examples. We observe from the first four columns that all the predictions of the model are meaningful, but the model is only incentivized when it matches the label. The task of CZSL is a multi label one and future datasets need to account for this. The last column shows some examples of suboptimal labels and wrong predictions. Right: We show good candidates for retrival on all three dataset and then we perform cross-dataset retrieval for a unseen composition across C-GQA and MIT-States.}
    \label{fig:qual}
\end{figure*}

\myparagraph{Graph architecture.}
\label{sec:graphablataion}
We ablate over the architecture of the graph at various depths from 2-8 layers to quantify the degree of knowledge propagation needed to achieve best performance. From Figure \ref{fig:gcnvsgcnii} we observe that a shallow architecture at 2 layers achieves the best AUC of 8.6 outperforming the deeper configuration. This is an established problem for the spectral graph convolution and is caused by laplacian smoothing across deeper layers\cite{li2018deeper}. 
To study if we are limited by a shallow representation, we use a more recent formulation of graph convolution named GCNII\cite{gcnii}. 
This method introduces a few key improvements like skip connections that remedy the laplacian smoothing problem. 
We see that while GCNII suffers less from the smoothing problem and maintains performance at deeper architectures, It only achieves an AUC of 7.2 for the best performing model. Since our graph is exploiting close relations between the states, objects and the compositions introduced by the dense connections for visual primitives, we are not held back by the shallow architecture. We advice future works to explore richer graphs that can facilitate deeper models.

\subsection{Qualitative results}
We show some qualitative results for the novel compositions with their top-3 predictions in Figure \ref{fig:qual} (left).
The first three columns present some instances where the top prediction matches the label. For MIT-States and C-GQA, we notice that the remaining two answers of the model contain information visually present in the image but not in the label highlighting a limitation of current CZSL benchmarks. Different groups of states like color, age, material etc. can represent different information for the same object and are thus not mutually exclusive. 

For example in column 4, row 1 the image of the cat consist of a size, surface texture and age all present in the label space of the dataset and the output of the model. However the label for this image only contains its surface texture.
This puts an upper limit on compositional class accuracy dependent on the number of groups associated with an object in the label space. Specifically, column 4 of Figure \ref{fig:qual} (left) counts as a wrong prediction but all of the top 3 predictions represent correct visual information for MIT-States and C-GQA. Unless the model learns the annotator bias, it can not achieve a perfect accuracy. 
Finally in column 5, we show some instances of sub-optimal and wrong labels. Specifically, the image in row 1 is entirely missing the thawed meat represented in the label, the image in row 2 can not sufficiently communicate the material information while the label in row 3 does not contain the dominant information in the image.

In Figure \ref{fig:qual} (right) we first show image retrieval results from seen and unseen compositions. We can see that for all three datasets our method returns the correct top images for the query. We then perform cross-dataset retrieval between MIT-States and C-GQA for an unseen composition. We show a representative image from the original dataset and the top-3 retrievals from the cross dataset. While the datasets have a distribution shift between them, we see that retrievals are still meaningful. On MIT-States 2/3 retrieved images match a Mossy pond while the 3rd image is a grass field confused with the query. Similar trend is observed for the model trained on C-GQA for retrieval of a puffy pizza. The model confuses the top retrieval with a casserole followed by two images of pizzas. Nevertheless, the cross dataset retrieval shows promise towards further generalization for future works.

\section{Conclusion}
We propose a novel graph formulation for Compositional Zero-shot learning in the challenging generalized zero-shot setting. Since our graph does not depend on external knowledge bases, it can readily be applied to a wide variety of compositional problems. 
By propagating knowledge in the graph against training images of seen compositions, we learn classifiers for all compositions end-to-end. 
Our graph also acts like a regularizer and allows us to learn image representations consistent with the compositional nature of the task.
We benchmark our method against various baselines on three datasets to establish a new state of the art in CZSL in all settings. We also highlight the limitations of existing methods and knowledge bases. We encourage future works to explore datasets with structured compositional relations and richer graphs that will allow for deeper graph models.

\myparagraph{\newline Acknowledgments}
Partially funded by ERC (853489 - DEXIM) and DFG (2064/1 – Project number 390727645).

{\small
\bibliographystyle{ieee_fullname}
\bibliography{egbib}
}

% \newpage
\section*{\Large Supplementary Material}
\appendix
\section{Creating C-GQA}
We introduced a new benchmark for Compositional Zero-shot Learning (CZSL) in the main manuscript. This benchmarks is based on the original GQA~\cite{gqa} dataset which is annotated with scene graphs where each bounding box is labelled with state-object or any other relations in the scene. For the creation of the benchmark, we only consider the bounding boxes with a single state-object relation to be consistent with existing works. Bounding boxes smaller than $112\times112$ are excluded because they are in half the input size of most feature extractors. %We consider only the bounding boxes with one state and object label.
From these bounding boxes, we collect the vocabulary of state and objects to remove overlapping concepts between state and object classes. We also merge plurals and synonyms using first wordnet~\cite{wordnet} and then manual checking. This yields the final vocabulary of 457 states and 870 objects.

We define a novel composition as a state-object pair not present in the training set. We now want to generate a validation and test set consisting of seen and novel compositions. We partition the testset of GQA randomly with respect to scene graphs into the validation and test sets of C-GQA. We further add 20 percent of GQA training scene graphs to the scene graphs that will constitute the validation and test set of CGQA. We divide these scene graphs with a probability of 0.45 and 0.55 respectively into validation and the test set. These numbers are chosen as the test set of C-GQA losses bounding boxes overlapping with the novel compositions in validation set. From the bounding boxes, we add the novel compositions in these graphs to the unseen set $\mathcal{Y}_{n-val}$ and $\mathcal{Y}_{n-test}$ respectively. However, the number of novel compositions is very small compared to the compositions in the training set represented by $\mathcal{Y}_s$. Therefore, we further divide the remaining compositions in validation randomly into $\mathcal{Y}_s$ and $\mathcal{Y}_{n-val}$. We then remove the unseen compositions of the validation set from the test set and divide the remaining compositions randomly into $\mathcal{Y}_s$ and $\mathcal{Y}_{n-test}$. Finally we remove the novel compositions $\mathcal{Y}_{n-val}$ and $\mathcal{Y}_{n-test}$ from $\mathcal{Y}_s$ and generate the images from the bounding boxes of the scene graph for the 3 sets.

This results in a training set consisting of 6963 pairs across 26k images; a validation set consists of 1173 seen and 1368 unseen pair across 7k images; and a testset consists of 1022 seen and 1047 unseen pairs across 5k images. In total C-GQA has a compositional space of over 9.3k compositional concepts making it the most extensive dataset for CZSL. With cleaner labels and a bigger label space, we hope this dataset is able to accelerate the research in the field.
%%%%%%%%%%%%%%%%%%%%%%%%%%%%%%%%%%%%%%%%%%%%%%%%%%

\section{Additional Experiments}

% \subsection{AUC at different k.}
% We reported the top1 AUC for the three datasets in the Table 2 of our main manuscript. We report top 1,2,3 AUC for the 3 datasets in table \ref{tab:auc} to allow direct comparison with older works that adopt this evaluation. We see that the trend from the main manuscript is consistent at different k. In particular on MIT-States, our model CGE achieves top3 AUC of 21.3 compared to 12.3 of the closest baseline Symnet continuing our 2$\times$ improvement. On UT-Zappos, CGE maintains its lead by achieving a top3 AUC of 77.5 compared to 69.8 of TMN. Finally on C-GQA, CGE again achieves a 2$\times$ improvement by achieving a top3 AUC of 6.4 compared to 3.3 of Symnet. 

%-------------------------------------------------------------------------

\subsection{End-to-end training with baselines}
We studied the impact of feature representations on the performance of the model in section 4 of the main paper. We showed that our model CGE benefits greatly from end to end training. Older baselines TMN and Symnet operate on a frozen ImageNet trained Resnet18 CNN backbone for feature extraction in their respective manuscripts. In this experiment, we train TMN and Symnet end to end (represented by EE) from the ImageNet- pretrained Resnet18~(the same with our CGE) and quantify if they are held back by the ImageNet representations on the validation set of MIT-States. 
%%%%%%%%%%%%%%%%%%%%%%%%
\begin{table}
\centering
\begin{tabular}{l|cc}
% \toprule
\textbf{Method} & \textbf{AUC} & \textbf{Best HM}   \\
\midrule
TMN EE \cite{tmn} & 2.9 & 13.0 \\
Symnet EE \cite{symnet} & 3.9 & 15.3 \\
CGE (Ours)      & \textbf{8.6} & \textbf{23.4} \\
% \bottomrule
\end{tabular}
\caption{End-to-End training results}\label{tab:endtoend}
\end{table}
%%%%%%%%%%%%%%%%%%%%%%%%%%%%%%%

We see from table \ref{tab:endtoend} that finetuning the CNN backbone  results in worse performance than in the original implementations as they are overfitting to the training set with an AUC of 2.9 for TMN\cite{tmn} and 3.9 for SymNet\cite{symnet}, while end-to-end training is beneficial for our CGE which achieves an AUC of 8.6 because of our graph regularization.

\begin{table}[]
\resizebox{\linewidth}{!}
{\begin{tabular}{l|cccccc}
\multirow{2}{*}{\textbf{Dataset}} & \multicolumn{6}{c}{\textbf{Embedding Model}}                \\
                                  & gl   & w2v           & ft   & gl+w2v & ft+gl & ft+w2v       \\
\midrule
MIT-States \cite{mitstates}                        & 6.4  & 6.4           & 6.5  & 6.6    & 6.6   & \textbf{6.8} \\
UT-Zappos \cite{utzappos1}                        & 38.6 & \textbf{38.7} & 37.5 & 37.0   & 36.2  & 38.1         \\
C-GQA (Ours)                            & 3.4  & \textbf{3.5}  & 3.2  & 3.4    & 3.3   & 3.5         
\end{tabular}}
\caption{\textbf{Ablation over embedding}: We use three popular word embedding models. (ft: Fasttext\cite{fasttext}, w2v: Word2Vec\cite{word2vec} and gl: Glove\cite{pennington2014glove})}
\label{tab:emb}
\end{table}

\subsection{Ablation over the GCN}
We reported ablation over various components of the Graph Convolutional Network (GCN) used in our model in the section 4.2 of the main manuscript. We ablate over the remaining components of the GCN. For these experiments, we use the fixed feature extractor version of our model $CGE_{ff}$ to quantify the improvements directly from the graph wrt to the word embeddings used for initialization and the learnable GCN configuration.

\textbf{Choice of embedding.} We test three popular word embedding models and the concatenation of their features for every word to study their impact on the performance of our model. We report the results in Table \ref{tab:emb}. We see that MIT-States benefits most from the concatenation of fasttext and word2vec models as these models are closely related to achieve a AUC of 6.8. While UT-Zappos and C-GQA achieve the best results with Word2Vec at 38.7 and 3.5 AUC respectively.

\begin{table}[]
\begin{subtable}{\linewidth}\centering
\begin{tabular}{cc|ccccc}
                         & \multicolumn{1}{l}{} & \multicolumn{5}{c}{\textbf{Num layers}}    \\
                            &                      & 2    & 4    & 6    & 8    & 10   \\
\cmidrule[1pt]{2-7}
                            & 1024                 & 6.53 & 6.13 & 5.58 & 5.07 & 4.33 \\
\multirow{5}{*}{\rotatebox[origin=c]{90}{\rlap{\footnotesize{\textbf{Hidden dim}}}}}                             
                            & 2056                 & 6.59 & 6.20 & 6.14 & 5.68 & 5.10 \\
                            & 4096                 & \textbf{6.80} & 6.30 & 6.20 & 5.83 & 4.95 \\
                            & 8184                 & 6.59 & 6.27 & 6.27 & 5.63 & 4.71
\end{tabular}
\subcaption{ Ablation over GCN}\label{tab:gcn}

\begin{tabular}{cc|ccccc}
                         & \multicolumn{1}{l}{} & \multicolumn{5}{c}{\textbf{Num layers}}    \\
                            &                      & 2    & 4    & 6    & 8    & 10   \\
\cmidrule[1pt]{2-7}
                            & 1024                 & 5.56 & 5.21 & 5.44 & 5.43 & 5.12 \\
\multirow{5}{*}{\rotatebox[origin=c]{90}{\rlap{\footnotesize{\textbf{Hidden dim}}}}}                             
                            & 2056                 & 6.00 & 6.10 & 6.00 & 5.92 & 5.84 \\
                            & 4096                 & 6.11 & 6.00 & 6.22 & 6.11 & 5.76 \\
                            & 8184                 & 6.54 & 6.14 & 6.00 & 5.61 & 5.32
\end{tabular}
\caption{ Ablation over GCNII}\label{tab:gcnii}
\end{subtable}
\caption{Ablation over the depth and hidden dimension of the GCN on CGE\textsubscript{ff}}
\end{table}

\textbf{Graph architecture.}
We ablate over the learnable architecture of GCN at different depth and hidden dimension on the validation set of MIT-States and report results in table \ref{tab:gcn}. We observe that increasing the hidden dimension is generally beneficial when we go from $1024$ to $4096$ as the performance increases from an AUC of 6.53 to 6.80. However, increasing the hidden dimension from $4096$ to $8192$ decreases the AUC to 6.59 at 2 layers of GCN.  Increasing the depth of the GCN network generally results in a decrease in performance across all hidden dimensions. In particular, at $4096$ the  AUC decreases from 6.80 AUC to 4.95. In order to validate if this is a consequence of laplacian smoothing we use a recent version of graph convolution called GCNII\cite{gcnii}. We see from table \ref{tab:gcnii} that the performance decrease across columns is less pronounced at different hidden dimensions for this model. However, the best AUC achieved at 6.54 is less than we achieved with the original GCN indicating that this version of graph convolution is less beneficial for our problem.

\end{document}